\documentclass{ecai2008}
\usepackage{times}
\usepackage{graphicx}
\usepackage{latexsym}
\usepackage[english]{babel}
\ecaisubmission 
\usepackage{amsthm}
\usepackage{epic,eepic,epsfig}
\usepackage{graphicx}
\usepackage{listings}
\usepackage{amsfonts}
\usepackage{subfigure}
\begin{document}
\newtheorem{lemma}{Lemma}
\newtheorem{observation}[lemma]{Observation}
\newtheorem{proposition}[lemma]{Proposition}
\newtheorem{definition}[lemma]{Definition}
\newtheorem{fact}[lemma]{Fact}
\newtheorem{corollary}[lemma]{Corollary}
\newtheorem{example}[lemma]{Example}

\author{Esben Rune Hansen\institute{IT-University of Copenhagen} \and
  S.~Srinivasa Rao\institute{MADALGO, Aarhus University, Denmark} \and
  Peter Tiedemann\institute{IT-University of Copenhagen} }

\title{Compressing Binary Decision Diagrams}

\maketitle
\bibliographystyle{ecai2008}

\begin{abstract}
  The paper introduces a new technique for compressing Binary
  Decision Diagrams in those cases where random access is not
  required.  Using this technique, compression and decompression can
  be done in linear time in the size of the BDD and compression will
  in many cases reduce the size of the BDD to 1-2 bits per node.

  Empirical results for our compression technique are presented,
  including comparisons with previously introduced techniques, showing
  that the new technique dominate on all tested instances.
\end{abstract}

\section{Introduction}\label{sec:introduction}
In this paper we introduce a technique for compressing binary decision
diagrams for those cases where random access to the compressed
representation is not needed. We will use the term \emph{offline} to
describe a BDD stored in such a manner that it no longer allows random
access to its structure without decompressing the BDD first. The two
primary areas in which decision diagrams are used in practice are
verification and configuration. In both of these areas it is sometimes
important to store binary decision diagrams offline using as little
space as possible. Primarily the need for such compression arises when
it is necessary to transmit binary decision diagrams across
communication channels with limited bandwidth. In the area of
verification this need arises for example when using a networked
cluster of computers to perform a distributed compilation of a binary
decision diagram \cite{distributed_verification}. A similar exchange of
BDD data takes place in \emph{distributed configuration} as described
in \cite{distributed_config}. In such approaches the fact that the
network bandwidth is much lower than the memory bandwidth can become a
major bottleneck as computers stall waiting to receive data to
process. Transmitting the binary decision diagrams in a compressed
representation can help alleviate this problem. The same problem
arises in standard interactive configuration. Consider for example a
web-based interactive configurator that uses BDDs to store the set of
valid configurations. It must either transmit the BDD storing the
configuration data to the customers computer or perform all
computations on the server leading to a network delay each time the
user makes a choice during the configuration. In order to reduce the
required bandwidth as well as lower the load time in the first case, it
is benefical to transmit the BDD in a compressed format.

The paper is organized as follows: In Section \ref{sec:preliminaries}
we present the neccessary notation and definitions. In Section
\ref{sec:compression} we present our compression scheme. Finally, in
Section \ref{sec:experiments} we show the compression attained by
applying our compression scheme on different instances.

\subsection{Related work}
Most of the work done on compressing decision diagrams aim to achieve
large reductions in size, while maintaining random access, by means of
better variable orderings \cite{fanin}\cite{exact} or modifications to
the decision diagram data structure \cite{embedded}. Such work is
mostly orthogonal to the aim of this paper, as the compression
strategy we present easily can be adapted for use with the variations
over basic BDDs.

The only previous work we are aware of for compressing BDDs for
offline storage is the work by Starkey and
Bryant\cite{bdd_media_compression} and the follow-up paper by Mateu
and Prades-Nebot\cite{bdd_image_compression} which both describes 
techniques for image compression using BDDs. The latter of these
includes a non-trivial encoding algorithm for storing the BDD
offline. Finally Kieffer et.al\cite{universal_bdd_compression} gives
theoretical results for using BDDs for general data compression
including a technique for storing BDDs. After presenting our own
technique, we present empirical results comparing the new encoder with
the encoders from \cite{bdd_image_compression} and
\cite{universal_bdd_compression}.

\section{Preliminaries}\label{sec:preliminaries}
\begin{definition}[Ordered Binary Decision Diagram]\cite{BRYANT-BDD}
  An ordered binary decision diagram on $n$ binary variables $X = \{
  x_1, \ldots, x_n \}$ is a layered directed acyclic graph $G(V,E)$
  with $n+1$ layers (some of which may be empty) and exactly one
  root. We use $l(u)$ to denote the layer in which the node $u$
  resides. In addition the following properties must be satisfied:
\begin{itemize}
\item If $|V| \ne 1$, there are exactly two nodes in layer
  $n+1$. These nodes have no outgoing edges and are denoted the
  1-terminal and the 0-terminal. If $|V| = 1$ the layer will either
  contain the 1-terminal or the 0-terminal.
\item All nodes in layer $1$ to $n$ have exactly two outgoing
  edges, denoted the \emph{low} and \emph{high} edge respectively.  We
  use $low(u)$ and $high(u)$ to denote the end-point of the low and
  high edge of $u$ respectively.
\item For any edge $(u,v) \in E$ it is the case that $l(u) < l(v)$
\end{itemize}
\end{definition}

We use $E_{low}$ and $E_{high}$ to denote the set of low and high
edges respectively. An edge $(u,v)$ such that $l(u) + 1 < l(v)$ is
called a long edge and is said to skip layer $l(u) + 1$ to
$l(v)-1$. The length of an edge $(u,v)$ is defined as $l(v) - l(u)$.

\begin{definition}[Reduced ordered Binary Decision Diagram]
  An ordered Binary Decision Diagram is called \emph{reduced} iff for
  any two distinct nodes $u,v$ it holds that $low(u) \not= low(v) \lor
  high(u) \not = high(v)$ and further that $high(u)\not = low(u)$ for
  all nodes $u$.
\end{definition}

In the rest of this paper we will assume for all BDDs we are
considering that they are ordered and reduced.

\begin{definition}[Solution to a BDD]
  A complete assignment $\rho$ to $X$ is a solution to an
  BDD $G(V,E)$ iff there exists a path $P$ from the root in $G$ to
  the 1-terminal such that for every assignment $(x_i, b) \in \rho$,
  where $b \in \{ low,high \}$, there exists an edge $(u,v)$ in $P$
  such that one of the following holds:
\begin{itemize}
\item[\textbf{-}] $l(u) < i < l(v)$
\item[\textbf{-}] $l(u) = i$ and $(u,v) \in E_b$
\end{itemize}
\end{definition}

\begin{figure}
\subfigure[]
{\includegraphics[scale=0.44]{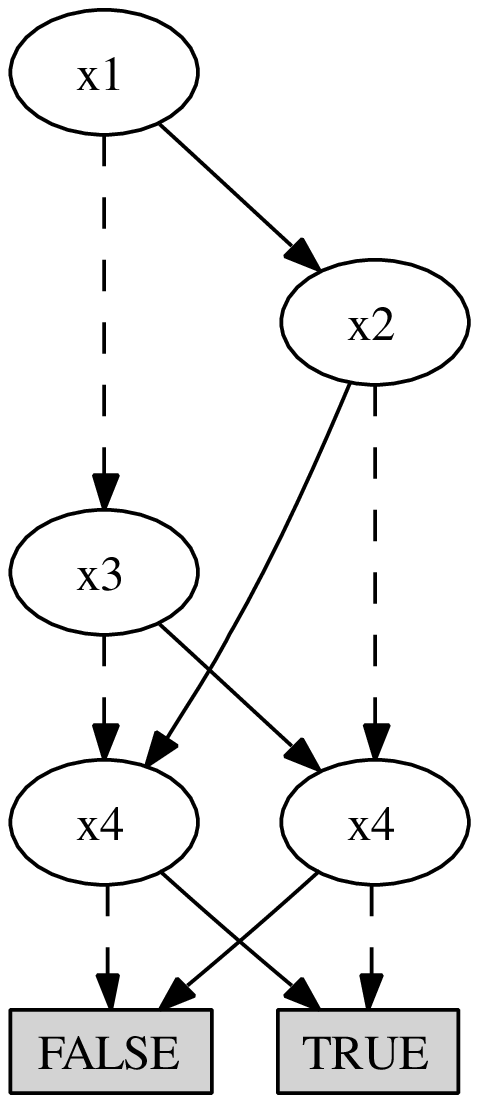}\label{fig:bdd}} \hfill
\subfigure[]
{\includegraphics[scale=0.44]{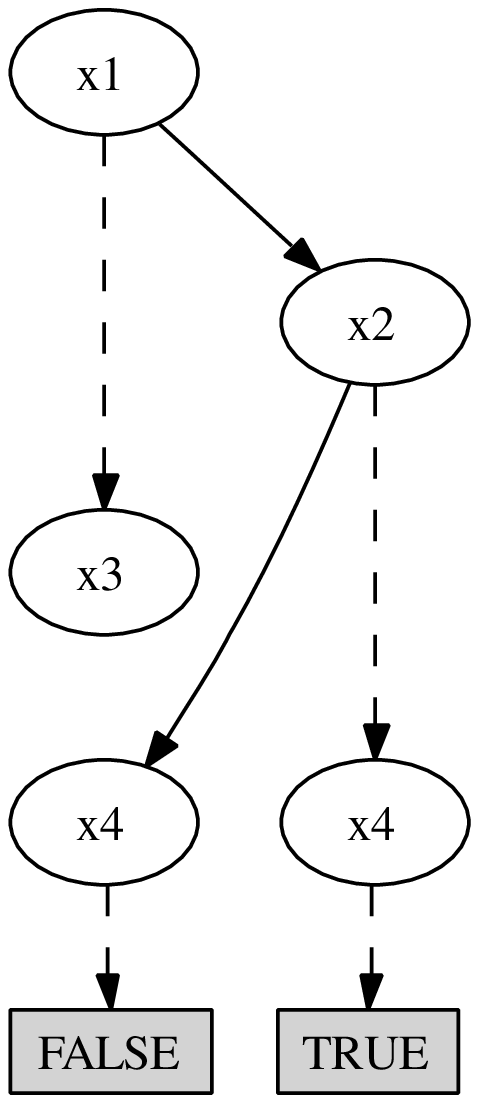}\label{fig:spanning1}} \hfill
\subfigure[]
{\includegraphics[scale=0.44]{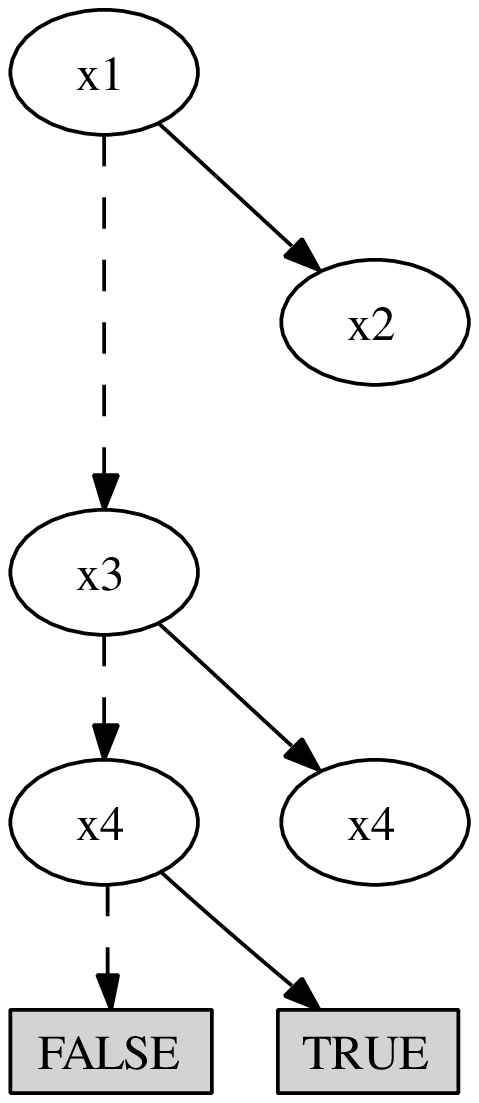}\label{fig:spanning2}} \hfill
\caption{From left to right: A BDD and two different spanning trees on the BDD. Solid and dashed edges corresponds to
\emph{high} and \emph{low} edges respectively}
\end{figure}

\begin{example}
  Figure \ref{fig:bdd} contains a BDD on the binary variables $X
  = \{x_1,x_2,x_3,x_4\}$ which the solutions
  $\{(1,1,1,1),$ $(1,1,0,1),$ $(1,0,1,0),$ $(1,0,0,0),$ $(0,1,1,0),$
  $(0,1,0,1),$ $(0,0,1,0),$ $(0,0,0,1)\}$. 
\end{example}

\begin{definition}[BFS order]\label{def:bfs_order}
  A BFS ordering $id_b: V \rightarrow \{1,\ldots,|V|\}$ of the nodes
  in a layered DAG $G(V,E)$ rooted in $r$ is the ordering of $V$ in
  the order they are visited by a BFS in the DAG starting at $r$ and
  traversing left edges prior to right edges. 
\end{definition}

\begin{definition}[Layer order]\label{def:layer_order}
  A layer ordering $id_l: V \rightarrow \{1,\ldots,|V|\}$ of the nodes
  in a layered DAG $G(V,E)$ rooted in $r$ is the ordering of $V$ layer
  by layer in increasing order of the layer. Nodes at the same layer
  are ordered in the order that they are visited by a DFS in the DAG
  starting at $r$ and traversing left edges prior to right edges.
\end{definition}
We refer to $id_b(v)$ and $id_l(v)$ by ``the BFS id of $v$'' and ``the
layer id of $v$'' respectively. Note that if all edges in a layered
DAG has the same length then the ordering $id_l$ and $id_b$ will be
the same.

In our compression scheme we will make use of the following well-known
fact:
\begin{lemma}\label{lemma:succinct}
Every binary tree can be unambiguously encoded using 2 bits pr. node.
\end{lemma}
To achieve such an encoding each node $v$ is encoded using two
bits. The first bit and the second bit is true iff $v$ contains a left
and a right child respectively. In order to make decoding possible the
order in which the children of already decoded nodes appear in the
encoded data must be known. This can for example be ensured by
encoding the nodes in a DFS or BFS order with either left-first or
right-first traversal. As an example, the encoding of the nodes of the
binary tree in Figure \ref{fig:spanning2} in BFS order is
$(11,11,00,11,00,00,00)$.

\section{The Compression technique}\label{sec:compression}
Our compression technique can be summarized by the following steps:
\begin{enumerate}
\item Build a spanning tree on the BDD (Section \ref{sub:spanning}).
\item Encode edges in the spanning tree, using Lemma \ref{lemma:succinct}
\item Encode by one bit the order in which the two terminals appear in
  the spanning tree.
\item Encode the length of the edges in the spanning tree where
  necessary (Section \ref{sub:layerinfo}).
\item Encode the edges that are not in the spanning tree (Section
  \ref{sub:nontreeedges}).
\item Compress the resulting data using standard compression
techniques.
\end{enumerate}

The decoder starts by reverting step (6) by decompressing the data. It
then recreates the binary tree (1-2), restores the correct layer of
each node (4), and restores the remaining missing edges (5). Below we
give the details of each step.

\subsection{Building the spanning tree on the BDD}\label{sub:spanning}
\begin{definition}[Spanning Tree]
  A spanning tree $G^T(V^T,E^T)$ on a BDD $G(V,E)$ is a subgraph of
  $G$, for which $V^T = V$, and any two vertices are connected by
  exactly one path of edges in $E^T$. An edge is called a \emph{tree
    edge} if it is contained in the spanning tree and a \emph{nontree
    edge} otherwise.
\end{definition}

The most obvious way to construct a spanning tree on a graph is to use
DFS or BFS. In the case of a rooted DAG one can obtain a spanning tree
by, for each node $v$ except the root, adding a single edge with
endpoint in $v$ to the set of tree edges. Two examples of spanning
trees for the BDD in Figure \ref{fig:bdd} are shown in Figure
\ref{fig:spanning1} and \ref{fig:spanning2}.

In our encoder we will construct a spanning tree containing as few
long edges as possible. Hence when a node $v$ in the BDD has multiple
parents $u_1,\ldots,u_k$ and we have to choose one of the edges
$(u_1,v),\ldots,(u_k,v)$ to add to the spanning tree, we will always
choose the shortest possible edge, that is an edge $(u_i,v)$ where
$l(u_i) \geq l(u_1),\ldots,l(u_k)$. Ties are broken by choosing the
edge $(u,v)$ with the smallest $id_l(u)$. Note that the resulting
spanning tree is uniquely defined regardless of which order we process
nodes in. Using this construction we achieve a spanning tree with a
minimal number of long edges. This can easily be seen by noting that
precisely one ingoing edge must be chosen for each node, and
additionally that the choice of one edge can never exclude an edge to
another node from consideration.


\begin{example}
  The spanning tree in Figure \ref{fig:spanning1} contains three long
  edges, whereas the spanning tree in Figure \ref{fig:spanning2} only
  contains one. The latter of these would be the one constructed by
  our encoder upon compressing the BDD in Figure \ref{fig:bdd}. The
  single long edge in figure \ref{fig:spanning2} has to be included in
  the tree as it is the only possible way for the spanning tree to
  include the node in layer 1.
\end{example}

\subsection{Encoding the lengths of the tree edges}\label{sub:layerinfo}
The spanning tree is stored as a binary tree where all edges have the
same length. Since some of the edges in the spanning tree may
correspond to long edges in the BDD, the binary tree itself may not be
sufficient to reconstruct the layer information of the nodes during
decoding. In order to enable the decoder to deduce the correct layer
we therefore encode the location and the length of each long edge that
is included in the spanning tree. The location of a long edge $(u,v)$
is uniquely specified by the BFS order of the end point of the edge,
that is $id_b(v)$.

When encoding the location of the long edges
$(u_1,v_1),\ldots,(u_k,v_k)$ we will, instead of outputting the
integers $id_b(v_1), \ldots id_b(v_k)$, output a bitvector of length
$|V|$ for which entries $id_b(v_1), \ldots ,id_b(v_k)$ are true and
all other entries are false. Though this encoding is likely to require
more bits than encoding by listing the integers, the bitvector will be
compressed very efficiently when the standard compression is applied
in the final phase.

\begin{figure}
\begin{center}
\includegraphics[scale=0.46]{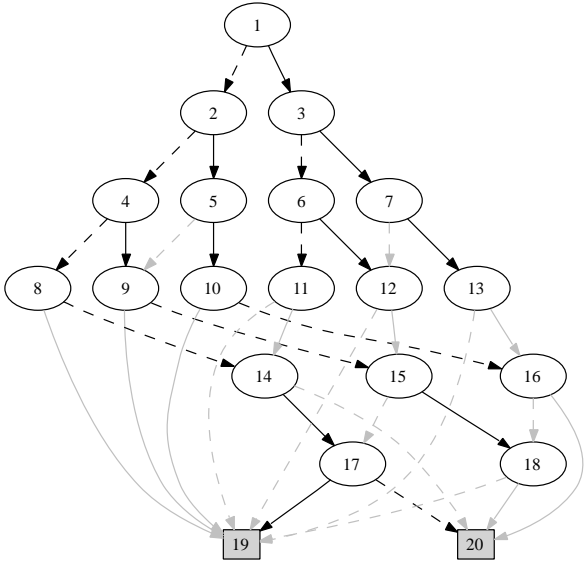}
\end{center}
\caption{ A spanning tree on a BDD. The black edges are tree edges and
  the gray edges are nontree edges. The nodes are labeled in layer
  order}
\label{fig:bdd+tree}
\end{figure}

\subsection{Encoding nontree edges}\label{sub:nontreeedges}
When the spanning tree and the layer information is encoded, we only
need to encode the nontree edges, that is, those edges in the BDD that
are not contained in the spanning tree.

We know that half of the edges in the BDD will be encoded as nontree
edges as it follows from the following observation:
\begin{observation}
  Let $G(V,E)$ be a BDD containing at least $3$ nodes. Then any
  spanning tree on $G$ will contain exactly $|E|/2+1$ edges
\end{observation}
\begin{proof}  
  By the assumption that $|V| > 2$, it holds for any BDD $G(V,E)$ that
  $|E| = 2(|V|-2)$, since all nodes in a BDD except the two terminals
  have two children. Further any tree with $|V|$ nodes contains
  $|V|-1$ edges, which equals $|E|/2 + 1$ edges.
\end{proof}

As an example the construction in Figure \ref{fig:bdd+tree} contains
19 tree edges and 17 nontree edges.

Since every node except the terminals in the BDD has two children and
we have the spanning tree available with restored layer information,
we know the start point of each of the nontree edges that has to be
added to the spanning tree in order to reconstruct the BDD. Hence if
we encode the nontree edges in some fixed order according to their
start point then we only need to encode the end point of each of the
nontree edges. We will call the endpoint of any nontree edge that has
yet to be encoded an \emph{incomplete child}. By $S$, we will denote
the sequence of incomplete children in the order in which their
parents appear in the layer ordering of the nodes, that is $S =
(v_1,\ldots, v_k)$ for the nontree edges
$(u_1,v_1),\ldots,(u_k,v_k)$. By $id_l(S)$ we denote the corresponding
sequence of layer ids, that is $id_l(S) =
(id_l(v_1),\ldots,id_l(v_k))$.

We will now describe three encodings of nontree edges in Section
\ref{sub:largeindegree}, Section \ref{sub:smallindegree} and Section
\ref{sub:longforward} which we will combine to encode all the
nontree edges.

\subsubsection{Incomplete children with large in-degree}\label{sub:largeindegree}
In order for a standard compression technique to successfully compress
the sequence $id_l(S)$ it is important that the symbols in the
sequence appears with high frequency. We note that nodes with
in-degree $d$ will appear $d-1$ times in the sequence of nontree
edges. Hence by applying standard compression we will be able to
efficiently compress those nontree children that have a high in-degree
if they are separated from the nodes that have a low
in-degree. Therefore we split the sequence of nodes appearing as
incomplete children in $S$ into two disjoint subsequences $H$ and $L$,
the first containing those incomplete children that have an in-degree
larger than a specified threshold in their order of appearence in $S$,
the latter containing the rest. For example if the threshold is 3 then
$H=(19,19,19,19,19,19,19,19,20,20,20,20)$ for the BDD in Figure
\ref{fig:bdd+tree}. Based on $H$ we construct the sequence of integers
$S^H$ on the sequence of nodes $v_1, \ldots ,v_{|V|}$ in $S$ by
encoding $v_i \in H$ as $id_l(v_i)$ and $v_i \in L$ as $0$. By the
encoded $0$s we indicate the incomplete children that are not among
the incomplete children with high in-degree. Note that all integers
appearing in $S^H$ occurs with a high frequency and therefore will
compress efficiently using standard compression techniques. The
remaining incomplete children $L$, we code separately, as described in
the next two sections.

\subsubsection{Incomplete children with small in-degree}\label{sub:smallindegree}
Using the above encoding, we are left with a sequence of nontree
children $L$, with very few repetitions. When encoding this sequence
we will exploit the fact that the sequence of integers in $id_l(L)$
will in most instances tend to be increasing. Below we argue why this
is the case.

\begin{enumerate}
\item The first reason is that any node $u$, with an outgoing edge of length
$k$ is naturally restricted to the children occurring in layer
$l(u)+k$, and therefore to the range of $id_l$ indices of the nodes at
layer $l(u)+k$. Since most edges in a BDD are usually rather short
(except those to the terminals), this leads to a natural increasing
progression in $id_l(L)$.

\item Secondly, when examining a set of layers in a BDD it is very
common to see disjoint substructures. For example, in Figure
\ref{fig:bdd+tree}, we have two disjoint substructures induced by the
nodes $2,4,5,8,9,10$ and $3,6,7,11,12,13$. Given two disjoint
substructures then for any given layer $l$ let $I^l_1$ and $I^l_2$ be
the sequences of layer indices of the nodes in that layer from each of
them respectively. Assume for convinience that $I^l_1$ contains the
smallest index. Then it is the case that all indices in $I^l_1$ are
strictly smaller than all indices in $I^l_2$, and furthermore the same
applies to $I^j_1$ and $I^j_2$ for any other layer $j$. Since the
incomplete children of a layer are encoded according to the order of
their parents, this means that $I^i_1$ will always appear in $id_l(L)$
before $I^{i}_2$ for any layer $i$, helping to ensure an increasing
tendency in $id_l(L)$.




\item Thirdly, some possible nontree edges cannot exist, since had
  they in fact existed they would have been included in the spanning
  tree. This constraint on
  the nontree edges is stated in the following observation:

\begin{observation}\label{obs:illegaledges}
  For every nontree edge $(u,v)$ it holds that $$v \not \in \{v' \mid
  \exists (u',v') \in E^T: l(u') = l(u) \land id_l(u) < id_l(u')\}$$ In other
  words: assume that there is a nontree edge $(u,v)$ for which a right
  sibling $u'$ of $u$ has the child $v'$ in the spanning tree, then $v \ne
  v'$.
\end{observation}
\begin{proof}
  If $v = v'$ then since $u$ and $u'$ are on the same layer, and since
  $id_l(u) < id_l(u')$ the spanning tree would contain $(u,v)$ rather
  than $(u',v)$, which contradicts that $(u,v)$ is a nontree
  edge. This is due to the fact that when the spanning tree is
  constructed by a traversal of the nodes in the BDD in layer order.
\end{proof}

\begin{example}
  In Figure \ref{fig:bdd+tree} an incomplete child of the
  node $5$ can neither be $11, 12$ or $13$, since in that case the
  corresponding edge would be a tree edge, which contradict that the
  child is incomplete. On the other hand an incomplete child of $7$
  can be any of the nodes $8,\ldots,12$ since $7$ is positioned to the right of
  all its siblings.
\end{example}

What follows from Observation \ref{obs:illegaledges} is, roughly
stated, that incomplete children with parents in the ``left part'' of
a layer are bound to have one of the smaller layer ids in the layer,
whereas the incomplete children with parents in the ``right part'' of
the of a layer can have any layer id occurring in the layer.
\end{enumerate}

\noindent As a conclusion of three the reasons mentioned above about
why we expect the sequence of incomplete children to tend towards being
increasing, and as we have observed the increasing trend of $id(L)$ in
the instances we have tested on, we choose to exploit this fact by
encoding the sequence $id_l(L)$ by delta coding:

\begin{definition}[Delta Coding]
Consider any sequence of integers $(i_1, \ldots i_k) \in \mathbb Z^k$
  for any $k \in \mathbb N$. We define the delta coding of $(i_1,
  \ldots i_k)$ by $\Delta(i_1, \ldots i_k) = (i_1, i_2-i_1, i_3-i_2,
  \ldots, i_k-i_{k-1})$
\end{definition}
For instance if $id_l(L) = (9,12,14,15,16,17,18,20)$, then
$\Delta(id_l(L)) = (9,3,2,1,1,1,1,2)$. Standard compression will be
able to compress the latter sequence much better than the former
sequence.

\subsubsection{Long forward edges}\label{sub:longforward}
Using the encoding of Section \ref{sub:smallindegree} nontree long
edges will often be expensive to encode since they have an incomplete
child with an id that is a lot larger than for the short edges. Hence
they will often result in large deltas in the delta coding. The
following approach is an alternative way of encoding some of the long
edges. This technique is therefore applied prior to the technique in
Section \ref{sub:smallindegree} and only the remaining edges will be
coded using delta-coding.

A nontree edge $(u,v)$ is a \emph{forward edge} if $u$ is an ancestor
of $v$ in the spanning tree. Consider any forward edge $(u,v)$ in the
graph with length $k$. This edge can be unambiguously encoded by
$id_l(v)$ and the length of the edge, as $u$ if the ancestor of $v$ at
layer $l(v)-k$.

In order to know which nodes that are endpoints of forward edges we
label each node $v$ by the number of forward edges ending in $v$.
This will be $0$ for most nodes and very seldom be more than $1$,
ensuring a good compression of the labelling. After this is done we
encode the length of the forward edges. If there are very few long
edges it might not be worth the effort to write the labelling on the
edges. Hence we set a threshold on the number of forward edges that it
needed in order to make the encoding of these edges useful. If the
threshold is not exceeded all long forward edges are instead encoded
as described in Section \ref{sub:smallindegree}.

\subsection{An example}
As a final part in the description of our compression technique, we
show how our technique would compress the BDD in Figure
\ref{fig:bdd+tree}.
\begin{example}
  Consider the encoding of the BDD in Figure \ref{fig:bdd+tree}. We
  first use Lemma \ref{lemma:succinct} to encode the spanning tree as
  the bitstring (comma separated only to ease readability)
  $11,1111,11011101, 101010000000,010100,1100,0000$. As there are no
  long edges in the spanning tree we do not need to encode layer
  information, we will output $0$ to denote that the total number of
  layer information that is to be added is $0$. If we suppose a
  threshold of $5$ in the encoding of nontree edges with high indegree
  only the node $19$ will be encoded as an incomplete child with high
  indegree. This will be encoded as
  $0\;0\;19\;19\;19\;19\;0\;19\;0\;19\;0\;0\;0\;0\;0\;19\;0$

We are now left with two long forward edges of length 2, namely
$(14,20)$ and $(16,20)$.  To encode them we first specify which of the
remaining nontree edges that are long forward edges by a bitvector
$0000010010$ and the length $2,2$. Finally we encode the remain
nontree edges by $9\;12\;14\;15\;16\;17\;18\;20$ which in delta coding
will be $9\;3\;2\;1\;1\;1\;1\;2$. 
\end{example}

\section{Experiments}\label{sec:experiments}
In this section we provide empirical results from compressing a large
set of BDDs from various sources using the new encoder described in
this paper and as well as the encoders from
\cite{bdd_image_compression} and \cite{universal_bdd_compression}. For
further comparison we also provide the results from a naive
encoder. The naive encoder outputs the size of each layer followed by
a list of children. This representation is very similar to the
in-memory representation of a BDD except that the layer information is
not stored for each node but rather implicitly using the layer sizes.

\subsection{Instances}
Many of the instances we show results for are taken from the
configuration library CLib \cite{clib}. As a BDD only allows binary
variables, additional steps must be taken in order to encode solutions
to problems containing variables with domains of size larger than
2. For each non-binary variable in a problem its customary to either
use a number of binary variables logarithmic in the size of the domain
of the variable and adjust the constraints accordingly or use one
variable for each domain value. These methods are known as
log-encoding\cite{yellowbook} and direct-encoding respectively. In the
instances we have tested with all those named with the suffix ``dir''
was compiled using direct encoding, while the remaining were build
using log-encoding. The instances fall into the following groups:

\paragraph{Product Configuration}
The instances in this group are all BDDs compiled for use with
standard interactive product configurators. For example the ``renault''
instance is a car configuration instance, and the others are various
PC configuration instances.

\paragraph{Power Supply Restoration}
These instances were compiled for use in configuring the restoration
of a power supply grid after a failure. As such they are also a type
of configuration instances.

\paragraph{Fault Trees}
These are instances built for use in reliability analysis using fault
trees.

\paragraph{Combinatorial}
The combinatorial group contains various ``toy'' chess problems of a
combinatorial nature. For example the classic problem of placing 8
queens on a chessboard without any piece being threatened is
represented by the instance ``8x8queen''. The ``5x27queens'' instance
models placing 5 queens on a 5x27 chess board.

\paragraph{Multipliers}
This group contains two BDDs both of which represent the value of the
middle bit in the output obtained by multiplying two groups of 10
input bits \cite{vlsi}. These are build mixing the input bits
(``mult-mix-10'') and separating the input bits (``mult-apart-10'').

\subsection{Post compression}
All the tested encoders create an encoding that is meant to be
subsequently compressed using standard entropy coding methods. In
\cite{bdd_image_compression} arithmetic coding is used while the
choice of entropy coding is not discussed in
\cite{universal_bdd_compression}. To avoid the empirical results being
affected by the choice of standard coding, we instead apply
LZMA\cite{lzma_sdk} to the output of all encoders to produce the final
encoding. Due to implementation details of this final compression
step, it is sometimes beneficial to produce the output that has to be
compressed on a byte level instead of a bit level. To ensure a fair
comparison the results stated for \cite{bdd_image_compression},
\cite{universal_bdd_compression} and the naive approach are obtained
by trying to output both on bit level and on byte level and
stating the best compression among the two results.  Our own encoder
was only tested outputting on a byte level.

\subsection{Conclusions}
From the empirical results shown in Figure \ref{fig:results} we can
immediately see that it is worthwhile to make use of a dedicated BDD
encoder, as the naive encoding, being only compressed by LZMA, is
outperformed with a factor of up to 20 on some instances. Furthermore
we can see that the encoder introduced in this paper is consistently
able to perform as well or better than the other encoders on all
tested instances. In particular the largest BDD in our test
(``complex-P3'') required about twice as much space when using either
of the two other dedicated encoders.

\paragraph{Instance properties}
For most of the instances included here it is the case that a very large
fraction (30\% to nearly 50\%) of the edges lead to the
zero-terminal. The exception to this are the multiplier instances,
``5x27queens'' and the ``rook'' instances. Slightly less expected is the
fact that it is quite rare for nodes other than the zero-terminal to
have a significant in-degree, this only occurring with any great
significance in ``5x27queens'' and to lesser extent in ``complex-P2''
and the multipliers. This means that in quite a few cases nearly all
of the non-tree edges are simply edges to the zero-terminal,
essentially turning $S^H$ into a bitvector, marking almost all the
edges as zero-terminal edges, allowing for very efficient
compression. This can be seen in the results where the ``5x27queens'',
the rook and the multiplier instances all turn out to compress less
efficiently. An additional important trend is that nodes which cannot
be reached by following a short edge from a parent are very rare,
meaning that our encoder in by far the most cases only need to provide
layer information for less than $1\%$ of the nodes, which is a
significant advantage over previous encoders.

\paragraph{Availability}
The Java source code used for these experiments (including a
command-line encoder and decoder for BDDs in the BuDDy \cite{buddy}
file format) is available along with all instances used in these
experiments at (URL removed for blind review).

\begin{figure}
\begin{center}
\footnotesize
\begin{tabular}{| c |  c |  c |  c | c | c|}
\hline
 \textbf{Name} & $|V|$ & \textbf{this paper}  & \cite{bdd_image_compression} &  \cite{universal_bdd_compression} & \textbf{Naive} \\
\hline
\multicolumn{6}{|c|}{\textbf{Product Configuration}} \\
\hline
renault          & 455798  & 0,90 & 126\% & 103\% & 402 \%\\
renault-dir      & 1392863 & 0,23 & 198\% & 214\% & 1352\%\\
pc-CP            & 16496   & 0,76 & 220\% & 209\% & 788 \%\\
pc               & 3467    & 2,19 & 224\% & 211\% & 436 \%\\
Big-PC           & 356696  & 0,38 & 334\% & 266\% & 1345\%\\
Big-PC-dir       & 1291600 & 0,17 & 260\% & 260\% & 2035\%\\
\hline
\multicolumn{6}{|c|}{\textbf{Power Supply Restoration}} \\
\hline
complex-P3       & 2812872 & 0,44 & 243\% & 202\% & 951 \% \\
complex-P2       & 163432  & 1,16 & 181\% & 167\% & 541 \%\\
1-6+22-32        & 20937   & 1,89 & 136\% & 154\% & 413 \%\\
1-6+22-32-dir    & 61944   & 0,99 & 135\% & 161\% & 606 \%\\

\hline
\multicolumn{6}{|c|}{\textbf{Fault Trees}} \\
\hline
isp9607          & 228706  & 0,63 & 389\% & 204\% & 873 \%\\
isp9605          & 4570    & 3,30 & 130\% & 145\% & 305 \%\\
chinese          & 3590    & 2,06 & 214\% & 160\% & 450 \%\\
\hline
\multicolumn{6}{|c|}{\textbf{Combinatorial}}\\
\hline
5x27queens     & 562764  & 4,33 & 108\% & 109\% & 204 \% \\
13x13rook        & 76808   & 3,56 & 210\% & 165\% & 311 \% \\
8x8rook          & 1339    & 6,03 & 140\% & 139\% & 277 \%\\
8x8queen-dir     & 2453    & 2,17 & 115\% & 178\% & 374 \%\\
8x8queen         &  879    & 4,29 & 114\% & 138\% & 332 \% \\
\hline
\multicolumn{6}{|c|}{ \textbf{Multipliers} }\\
\hline
mult-mix-10      & 42468   & 9,92*& 114\% & 107\% & 169 \%\\
mult-apart-10    & 31260   & 8,07*& 120\% & 124\% & 202 \%\\
\hline
\end{tabular}
\end{center}
\caption{The above table shows the name and size (in nodes) of each of
  the instances tested. The result of the new encoder, in bits per
  node, is then showed, followed by the relative results of the rest
  of the encoders. The * indicates results obtained from our encoder
  when delta coding is not used in the encoding of nontree edges.}
\label{fig:results}
\end{figure}
\bibliography{main}
\end{document}